%% file: SparseHash.tex
\documentclass{article}

\usepackage{graphicx} 
\usepackage{subfigure}

\usepackage{natbib}

\usepackage{amsfonts}
\usepackage{amsmath}
\usepackage{amssymb}

\usepackage{overpic}
\usepackage{bm}

\usepackage[accepted]{icml2014}
\usepackage{epsfig}

\usepackage{multirow}

\usepackage{flushend}

\usepackage{subfigure}
\usepackage{overpic}

\newcommand{\bb}[1]{\boldsymbol{\mathrm{#1}}}

\def\Tr{\mathrm{T}}

\newcommand{\EE}{\mathbb{E}}
\newcommand{\RR}{\mathbb{R}}
\newcommand{\HH}{\mathbb{H}}

\newcommand{\xx}{\bb{x}}

\newcommand{\sign}{\mathrm{sign}}

\newcommand{\aaa}{\bb{a}}

\newcommand{\Pp}{\bb{P}}

\usepackage[pagebackref=true,breaklinks=true,letterpaper=true,colorlinks,bookmarks=false]{hyperref}

\icmltitlerunning{Sparse similarity-preserving hashing}

\begin{document}


\twocolumn[

\icmltitle{Sparse similarity-preserving hashing}

\icmlauthor{Jonathan Masci
    }{jonathan@idsia.ch}
\icmlauthor{Alex M. Bronstein}{bron@eng.tau.ac.il}
\icmlauthor{Michael M. Bronstein}{michael.bronstein@usi.ch}
\icmlauthor{Pablo Sprechmann}{pablo.sprechmann@duke.edu}
\icmlauthor{Guillermo Sapiro}{guillermo.sapiro@duke.edu}


\icmlkeywords{...}

\vskip 0.3in
]


\begin{abstract}

In recent years, a lot of attention has been devoted to efficient nearest neighbor search by means of similarity-preserving hashing. 
One of the plights of existing hashing techniques is the intrinsic trade-off between performance and computational complexity: while longer hash codes allow for lower false positive rates, it is very difficult to increase the embedding dimensionality without incurring in very high false negatives rates or prohibiting computational costs.
In this paper, we propose a way to overcome this limitation by enforcing the hash codes to be {\em sparse}.
Sparse high-dimensional codes enjoy from the low false positive rates typical of long hashes, while keeping the false negative rates similar to those of a shorter dense hashing scheme with equal number of degrees of freedom. 
%
%
We use a tailored feed-forward neural network for the hashing function. Extensive experimental evaluation involving visual and multi-modal data shows the benefits of the proposed method. 

\end{abstract}

\input{intro.tex}
\input{background.tex}
\input{sparsity.tex}

\input{implementation.tex}

\input{results.tex}

\input{conclusion.tex}


{\small
\bibliographystyle{icml2014}
\bibliography{mmhash}
}

\newpage
\section*{}
\newpage
\section*{Supplementary Material for Sparse similarity-preserving hashing}
\input{supplementary}

\end{document}

%% file: intro.tex
\section{Introduction}


Efficient computation of similarity between entries in large-scale databases has attracted increasing interest, given the explosive growth of data that has to be collected, processed, stored, and searched for.
%
This problem arises naturally in applications such as image-based retrieval, ranking, classification, detection, tracking, and registration. In all these problems, given a query object (usually represented as a feature vector), one has to determine the closest entries (nearest neighbors) in a large (or huge) database.
Since the notion of similarity of (for example) visual objects is rather elusive and cannot be measured explicitly, one often resorts to machine learning techniques that allow constructing similarity from examples of data. Such methods are generally referred to as {\em similarity} or {\em metric learning}.

%
Traditionally, similarity learning methods can be divided into unsupervised and supervised, with the former relying on the data only without using any side information.
PCA-type methods \cite{scholkopf1997kernel} use global structure of the data, while manifold learning techniques such as locally linear embedding \cite{roweis2000nonlinear}, eigenmaps \cite{belkin2003laplacian}, and diffusion maps \cite{coifman2006diffusion} consider data as a low-dimensional manifold and use its local intrinsic structure to represent similarity.
%
%
Supervised methods assume that additional information, such as class labels  \cite{johnson2002applied,mika1999fisher,weinberger2009distance,xing2002distance}: distances, 
similar and dissimilar pairs \cite{davis2007information}, or order relations  \cite{mcfee2009partial,shen2009positive}, is provided together with the data examples. 
%
Many similarity learning methods use some representation of the distance, e.g., in the form of a parametric embedding from the original data space to some target space. In the simplest case, such an embedding is a linear projection acting as dimensionality reduction, and the metric of the target space is Euclidean or Mahalanobis distance \cite{shen2009positive,weinberger2009distance}.


More recently, motivated by the need for efficient techniques for big data, there has been an increased interest in similarity learning methods based on embedding the data in spaces of binary codes with the Hamming metric \cite{gong2012angular,gong2011iterative,kulis2009learning,liu2012supervised,norouzi2012hamming,norouzi2011minimal,wang2010sequential}. Such an embedding can be considered as a hashing function acting on the data trying to preserve some underlying similarity.
Notable examples of the unsupervised setting of this problem include {\em locality sensitive hashing} (LSH) \cite{gionis1999similarity} and {\em spectral hashing} \cite{weiss2008spectral,liu2011hashing}, which try to approximate some trusted standard similarity such as the Jaccard index or the cosine distance. Similarly, Yagnik et al. (\citeyear{yagnik2011power}) proposed computing ordinal embeddings based on partial order
statistics such that Hamming distance in the resulting space closely correlates
with rank similarity measures. Unsupervised methods cannot be used to learn semantic similarities given by example data. 
Shakhnarovich et al. (\citeyear{shakhnarovich2003fast}) proposed to construct optimal LSH-like {\em similarity-sensitive hashes} (SSH) for data with given binary similarity function using boosting, considering each dimension of the hashing function as a weak classifier. In the same setting, a simple method based on eigendecomposition of covariance matrices of positive and negative samples was proposed by \cite{strecha2010ldahash}.
Masci et al. (\citeyear{masci2011descriptor}) posed the problem as a neural network learning.
Hashing methods have been used successfully in various vision applications such large-scale retrieval \cite{torralba2008small}, feature descriptor learning \cite{strecha2010ldahash,masci2011descriptor}, image matching \cite{korman2011coherency} and alignment \cite{bronstein2010data}.

%
%
%
%

The appealing property of such similarity-preserving hashing methods is the compactness of the representation and the low complexity involved in distance computation: finding similar objects is done through determining hash collisions (i.e. looking for nearest neighbors in Hamming metric balls of radius zero), with complexity practically constant in the database size.
In practice, however, most methods consider nearest neighbors lying at larger  than zero radii, which then cannot be done as efficiently.
The reason behind this is the 
difficulty of simple 
hash functions (typically low-dimensional linear projections) to achieve simultaneously high precision and recall by only requiring hash collisions.
%

{\bf Main contributions.}
In this paper, we propose to introduce structure into the binary representation at the expense of its length, an idea
that has been shown spectacularly powerful and led to numerous applications of sparse redundant representation and compressed sensing
techniques. 
We introduce a \emph{sparse similarity-preserving hashing} technique, SparseHash, and show a substantial evidence of its superior recall 
at precision comparable 
to that of state-of-the-art methods, on top of its intrinsic computational benefits. 
To the best of our knowledge, this is the first time sparse structure are employed in similarity-preserving hashing.
We also show that the proposed sparse hashing technique can be thought of as a feed-forward neural network, whose architecture is motivated by the iterative shrinkage algorithms used for sparse representation pursuit \cite{daubechies2004iterative}. The network is trained using stochastic gradient, scalable to very large training sets.
Finally, we present an extension of SparseHash to multimodal data, allowing its use in multi-modal and cross-modality retrieval tasks.

%% file: background.tex
\section{Background}
\label{sec:formulation}

Let $X \subseteq \RR^n$ be the data (or feature) space with a binary similarity function $s_X : X\times X \rightarrow \{-1,+1\}$. In some cases, the similarity function can be obtained by thresholding some trusted metric on $X$ such as the $\ell_2$ metric; in other cases, the data form a (typically, latent) low-dimensional manifold, whose geodesic metric is more meaningful than that of the embedding Euclidean space. In yet other cases, $s_X$ represent a semantic rather than geometric notion of similarity, and may thus violate metric properties. 
It is customary to partition $X \times X$ into similar pairs of points ({\em positives})
$\mathcal{P} = \{ (\xx,\xx') : s_{X}(\xx,\xx') = +1\}$, and dissimilar pairs of points ({\em negatives})  $\mathcal{N} = \{ (\xx,\xx') : s_{X}(\xx,\xx') = -1\}$.

{\em Similarity-preserving hashing} is the problem of representing the data from the space $X$ in the space $\HH^m = \{ \pm 1\}^m$ of $m$-dimensional binary vectors with the Hamming metric $d_{\HH^m}(a, b) =  \frac{m}{2} - \frac{1}{2} \sum_{i=1}^m a_i b_i$ by means of an embedding $\xi: X \rightarrow \HH^m$ that preserves the original similarity relation, in the sense that there exist two radii, $r < R$ such that with high probability $d_{\HH^m} |_{\mathcal{P}} \le r$ and $d_{\HH^m} |_{\mathcal{N}} \ge R$.

In practice, the similarity $s_X$ is frequently unknown and hard to model, however, it is possible to sample it on some subset of the data. 
In this setting, the problem of similarity-preserving hashing boils down to finding an embedding $\xi$ minimizing the aggregate of false positive and false negative rates,
\begin{eqnarray}
\label{eq:sshloss}
\min_{\xi} \,\,  \EE \{ d_{\HH^m} \circ (\xi \times \xi) | \mathcal {P} \} - \EE \{ d_{\HH^m} \circ (\xi \times \xi) | \mathcal {N} \}.
\end{eqnarray}
Problem~(\ref{eq:sshloss}) is highly non-linear and non-convex.
We list below several methods for its optimization.

{\bf Similarity-sensitive hashing (SSH). }
Shakhnarovich et al. (\citeyear{shakhnarovich2003fast}) studied a particular setting of problem~(\ref{eq:sshloss}) with embedding of the form $\xi(\xx) = \sign(\Pp \xx + \aaa)$, where $\Pp$ is an $m\times n$ projection matrix and $\aaa$ is an $m\times 1$ bias vector, and proposed the SSH algorithm constructing the dimensions of $\xi$ one-by-one using boosting.
%
The expectations in~(\ref{eq:sshloss}) are weighted, where stronger weights are given to misclassified pairs from previous iteration.

{\bf Diff-hash (DH). }
Strecha et al. (\citeyear{strecha2010ldahash}) linearized the embedding $\xi(\xx) = \sign(\Pp \xx + \aaa)$ to $\hat{\xi}(\xx) = \Pp \xx + \aaa$, observing that in this case~(\ref{eq:sshloss}) can be written as
\begin{eqnarray}
\min_{\Pp^\Tr\Pp = \bb{I}} \mathrm{tr}(\Pp (\bb{\Sigma_+} - \bb{\Sigma_-} )\Pp^\Tr),
\label{eq:diffhash}
\end{eqnarray}
where $\bb{\Sigma_+}, \bb{\Sigma_-}$ are the covariance matrices of the differences of positive and negative samples, respectively. Solving~(\ref{eq:diffhash}) w.r.t. to the projection matrix $\Pp$ amounts to finding the smallest eigenvectors of the covariance difference matrix $\bb{\Sigma_+} - \bb{\Sigma_-}$. 
The vector $\aaa$ is found separately, independently for each dimension.

{\bf Neural network hashing (NN-hash). } Masci et al. (\citeyear{masci2011descriptor}) realized the function $\xi(\xx) = \mathrm{sign}(\Pp \xx + \aaa)$ as a single-layer neural network with $\tanh(\beta x) \approx \mathrm{sign}(x)$ activation function, where the coefficients $p_{ij}, a_i$ act as the layer weights and bias, respectively.
Coupling two such networks with identical parameters in a so-called {\em siamese architecture} \cite{hadsell-chopra-lecun-06,Taylor11}, one can represent the loss~(\ref{eq:sshloss}) as
\begin{eqnarray}
\vspace{-2mm}
\label{eq:mmnnhash-loss-X}
L(\Pp,\aaa)  \hspace{-2mm}& = & \hspace{-2mm}\frac{1}{2|\mathcal{P}|} \sum_{(\xx,\xx') \in \mathcal{P}} \| {\xi}(\xx) - {\xi}(\xx') \|_{2}^{2}   \\
			 \hspace{-2mm}   &  + & \hspace{-2mm}\frac{1}{2|\mathcal{N}|}  \sum_{(\xx,\xx') \in \mathcal{N}} \max\{0, M - \| {\xi}(\xx) - {\xi}(\xx') \|_{2} \}^{2}  \nonumber
\vspace{-3mm}
\end{eqnarray}
The second term in~(\ref{eq:mmnnhash-loss-X}) is a {\em hinge-loss} providing robustness to outliers and producing a mapping for which negatives are pulled $M$-apart.
%

Finding the network parameters $\Pp, \aaa$ minimizing loss function~(\ref{eq:mmnnhash-loss-X}) is done using standard NN learning techniques, e.g. the back-propagation algorithm \cite{LeCun:85}.
Compared to SSH and DH, the NN-hash attempts to solve the full non-linear problem rather than using often suboptimal solutions of the relaxed linearized or separable problem such as~(\ref{eq:diffhash}).
%

%
%

%% file: sparsity.tex
\section{Sparse similarity-preserving hashing}
\label{sec:sparsity}

The selection of the number of bits $m$ and the rejection Hamming radius $r$ in
a similarity-preserving hash has an important influence on the tradeoff between precision and recall.
%
%
%
The increase of $m$ increases the precision, as a higher-dimensional embedding space allows representing more complicated decision boundaries. At the same time, with the increase of $m$, the relative volume of the ball $B_r$ containing the positives decays exponentially fast, a phenomenon known as the curse of dimensionality, resulting in a rapid decrease of the recall.
This is a well-documented phenomenon that affects all hashing techniques \cite{grauman2013learning}.
For instance, in the context of LSH, it can be shown that the collision probability 
between two points decreases exponentially with the code-length \cite{goemans1995improved}.
Furthermore, increasing $m$ slows down the retrieval. 

The low recall typical to long codes can be improved by increasing the rejection radius $r$. However, this comes at the expense of increased query time, since the search complexity directly depends on the rejection radius $r$. 
For $r=0$ (collision), a look-up table (LUT) is used: the query code is fed into the LUT, containing all entries in the database having the same code. The complexity is $\mathcal{O}(m)$, independent of the database size $N$, but often with a large constant. 
For small $r$ (partial collision), the search is done as for $r=0$ using perturbation of the query: at most $r$ bits of the query are changed, and then it is fed into the LUT. The final result is the union of all the retrieved results. Complexity in this case is $\mathcal{O}({r \choose m})$. 
Finally, for large radii it is often cheaper in practice to use exhaustive search with complexity $\mathcal{O}(N)$ (for typical code lengths and database sizes used in vision applications, using $r>3$ is slower than brute-force search \cite{grauman2013learning}).   
Consequently, practical retrieval based on similarity-preserving hashing schemes suffers from a fundamental limitation of the precision-recall-speed tradeoff: one has to choose between fast retrieval (small $r$ and $m$, resulting in low recall), high recall (large $r$, small $m$, slow retrieval), or high precision (large $m$, small recall, and slow retrieval). 

\begin{figure*}[t!]
\centering
\includegraphics[width=.9\linewidth]{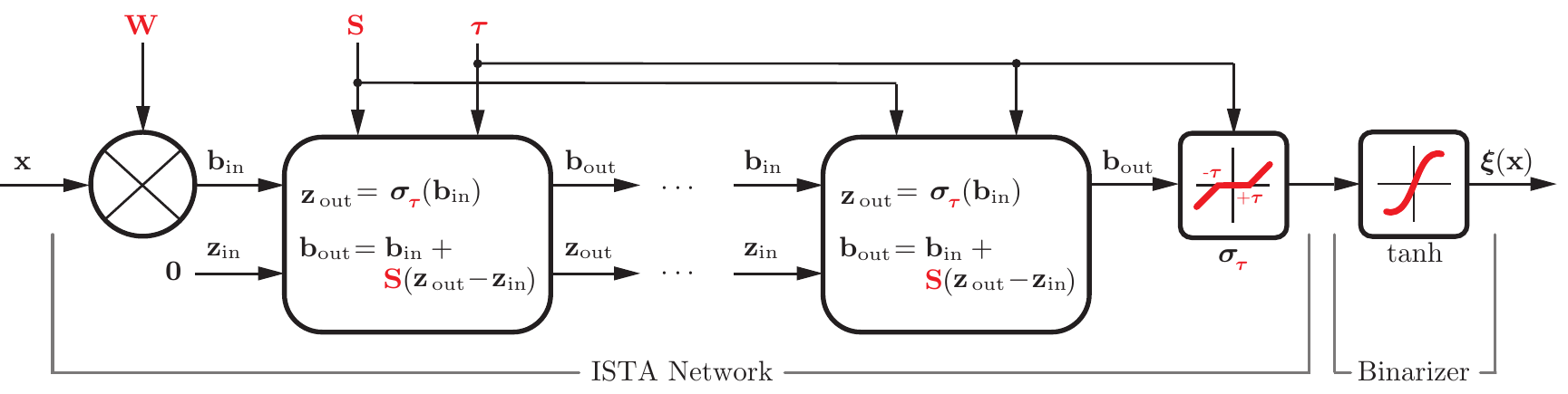} \vspace{-2mm}
\caption{\label{fig:ista}
Schematic representation of our ISTA-type network which we use to realize sparse hashing. Two such networks, with the same parametrization, are used.}
\vspace{-5mm}
\end{figure*}

The key idea of this paper is to control the exploding volume of the embedding space by introducing structure into the binary code. While different types of structure can be considered in principle, we limit our attention to \emph{sparse} hash codes. 
A number of recent studies has demonstrated that sparse over-complete representations have several theoretical and practical advantages when modeling compressible data leading to state-of-the-art results in many applications in computer vision and machine learning. 
%
%
%
We argue, and show experimentally in Section~\ref{sec:results}, that compared to its ``dense'' counterpart, 
an $m$-bit $k$-sparse similarity-preserving hash can enjoy from the high precision typical for long hashes, while having higher recall roughly comparable to that of a dense hashing scheme with $\log_2{m \choose k} = \mathcal{O}(k \log_2 m)$ bits (which has the same number of degrees of freedom of the $m$-bit sparse hash).

{\bf SparseHash. }
In order to achieve sparsity, a regularization needs to be incorporated into problem (\ref{eq:sshloss})
so that the obtained embedding $\xi$ will produce codes having only a small number of non-zero elements.
In this work we employ an $\ell_1$-norm regularization, extensively used in the compressed sensing literature to promote sparsity.
Specifically, the loss considered in the minimization of the proposed SparseHash framework is given by the average of 
\begin{eqnarray}
\label{eq:sh_loss}
\lefteqn{L(\xi(\bb{x}),\xi(\bb{x}'),s_{X})  =  s_{X}(\xx,\xx') \|\xi(\bb{x}) - \xi(\bb{x}')\|_1} \nonumber \\
&& + \frac{\lambda}{2} (1-s_{X}(\xx,\xx')) \max\{0, M - \|\xi(\bb{x}) - \xi(\bb{x}')\|_1 \}^2   \nonumber\\
&& + \alpha (\|\xi(\bb{x})\|_1 + \|\xi(\bb{x}')\|_1), 
\end{eqnarray}
over the training set, where $s_{X}$ is the groundtruth similarity function (1: similar, 0: dissimilar), $\alpha$ is a parameter controlling the level of sparsity, $\lambda$ is a weighting parameter governing the false positive and negative rate tradeoff, and $M$ is a margin.

With the new loss function given in (\ref{eq:sh_loss}), solving (\ref{eq:sshloss}) will produce a sparse embedding
that minimizes the aggregate of false positive and false negative rates for a \emph{given} parametrization. Now the question is what parametrized family of embedding functions would lead to the best sparse similarity-preserving hashing codes?
While there is no absolute answer to this question, recent approaches aimed at finding \emph{fast} approximations of sparse codes
have shed some light on this issue from a practical perspective \cite{LecunNN, pami2012}, and we use this same criterion for our proposed framework. 

Gregor and LeCun (\citeyear{LecunNN}) proposed tailored feed-forward architectures capable of producing highly accurate approximations of the true sparse codes.
These architectures were designed to mimic the iterations of successful first order optimization algorithms such as the iterative thresholding algorithm (ISTA) \cite{daubechies2004iterative}.
The close relation between the iterative solvers and the network architectures plays a fundamental role in the quality of the approximation.
This particular design of the encoder architecture was shown to lead to considerably better approximations than other of-the-shelf  feed-forward neural networks \cite{LecunNN}.
 These ideas can be generalized to many different uses of sparse coding, 
 in particular, they can be very effective in discriminative scenarios, performing similarly or better than exact algorithms at a small fraction of the computational cost \cite{pami2012}. 
These architectures are flexible enough for approximating
 the sparse hash in \eqref{eq:sh_loss}.\footnote{The hash codes produced by the proposed architecture can only be made on {\em average} $k$-sparse by tuning the parameter $\alpha$. In order to  guarantee that the codes contain no more than $k$ non-zeros, one can resort to the CoD encoders, derived from the coordinate descent pursuit algorithm \cite{LecunNN}, wherein $k$ is upper-bounded by the number of network layers. We stress that in our application the exact sparsity is not important, since we get the same qualitative behavior.   }


%% file: implementation.tex


{\bf Implementation. } 
We implement SparseHash by coupling two ISTA-type networks, sharing the same set of parameters as the ones described in \cite{LecunNN,pami2012}, and trained using the loss  (\ref{eq:sh_loss}). The architecture of an ISTA-type network (Figure~\ref{fig:ista}) can also be seen as a recurrent network with a soft threshold activation function.
%
A conventional ISTA network designed to obtain sparse representations with fixed complexity has continuous output units.
We follow the approach of \cite{masci2011descriptor} to obtain binary codes by adding a $\tanh$ activation function.
Such smooth approximation of the binary outputs is also similar to the logistic function used by KSH \cite{liu2012supervised}.
We initialize $\mathbf{W}$ with a unit length normalized random subset of the training vectors, $\mathbf{S}$ as in the original ISTA algorithm considering $\mathbf{W}^\Tr$ as dictionary and the thresholds $\bb{\tau}$ with zeros.
The shrinkage activation is defined as
$\sigma(\bb{x}, \bb{\tau}) = \max(0, |\bb{x}| - \bb{\tau}) \sign(\bb{x})$.

The application of the learned hash function to a new data sample
involves few matrix multiplications and the computation of the element-wise soft thresholding,
and is on par with the fastest methods available, such as  
SSH \cite{shakhnarovich2003fast}, DH \cite{strecha2010ldahash}, and AGH \cite{liu2011hashing}.

\section{Multi-modal SparseHash}

\label{subsec:mm}

In modern retrieval applications, a single object is often represented by more than one data modality. For example, images are frequently accompanied by textual tags, and video by an audio track. The need to search multi-modal data requires comparing objects incommensurable in their original representation. 
Similarity-preserving hashing can address this need by mapping the modalities into a common space, thus making them comparable in terms of a single similarity function, such as the Hamming metric.
For simplicity, we will henceforth limit our discussion to two modalities, though the presented ideas can be straightforwardly generalized to any number of modalities. 

We assume the data comes from two distinct data spaces $X$ and $Y$, equipped with \emph{intra-modality} similarity functions $s_X$ and $s_Y$, respectively.
We furthermore assume the existence of an \emph{inter-modality} similarity $s_{XY}$. Typically, examples of similar and dissimilar objects across modalities are more expensive to obtain compared to their intra-modality counterparts.
We construct two embeddings  $\xi: X \rightarrow \HH^m$ and $\eta: Y \rightarrow \HH^m$, in such way that the Hamming metric preserves the similarity relations of the modalities. We distinguish between a \emph{cross-modal} similarity-preserving hashing, preserving only the intra-modality similarity 
\cite{bronstein2010data}, and
the full \emph{multi-modal} setting, also preserving inter-modal similarities. 

Our sparse similarity-preserving hashing technique can be generalized to both settings. We construct an independent SparseHash network for each modality, and train them by minimizing an aggregate loss of the form
\begin{eqnarray}
L_{\textrm{MM}} &=& \mu_1 L(\xi(\bb{x}),\xi(\bb{x'}),s_{X}(\bb{x},\bb{x}')  )  + \hspace{-5mm}\\
&& \hspace{-15mm} \mu_2 L(\eta(\bb{y}),\eta(\bb{y}'),s_{Y}(\bb{y},\bb{y}')  ) + L(\xi(\bb{x}),\eta(\bb{y}),s_{XY}(\bb{x},\bb{y}))
,\nonumber
\end{eqnarray}
with respect to the parameters of the networks. The parameters $\mu_1$ and $\mu_2$ control the relative importance of the intra-modality similarity, and are set to zero in the \emph{cross-modal} regime. We refer to the networks constructed this way as MM-SparseHash.

%% file: results.tex
\input{res_cifar.tex}

\begin{figure}[t]
\centering
\includegraphics[width=1\linewidth]{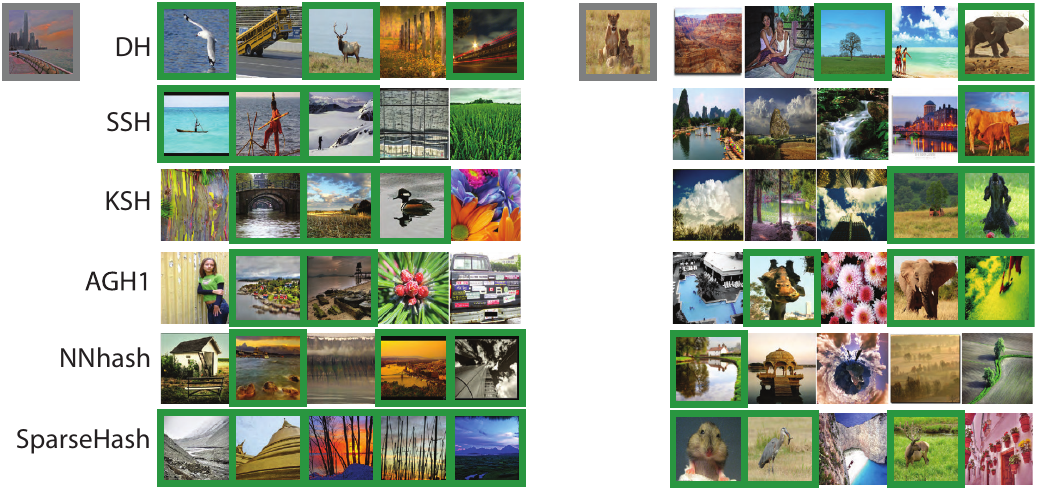}
\caption{\label{fig:nn-nus-5nn}
Five nearest neighbors retrieved by different hashing methods for two different queries (marked in gray) in the NUS dataset. 
Correct matches are marked in green. Note that in NUS dataset each image has multiple class labels, and our groundtruth positives are defined as at least one class label in common. }\vspace{-4mm}
\end{figure}

\section{Experimental results}
\label{sec:results}
We compare SparseHash to several state-of-the-art supervised and semi-supervised hashing methods: DH~\cite{strecha2010ldahash}, SSH~\cite{shakhnarovich2003fast}, AGH~\cite{liu2011hashing}, KSH~\cite{liu2012supervised}, and NNhash~\cite{masci2011descriptor}, using  
%
codes provided by the authors. 
%
%
%
For SparseHash, we use fully online training via stochastic gradient descent with annealed learning rate and momentum, fixing 
the maximum number of epochs to 250. A single layer ISTA net is used in all experiments.
All dense hash methods achieve an average sparsity of about $50\%$ per sample whereas SparseHash
achieves much sparser and structured codes; i.e. $6\%$ sparsity on CIFAR10 with hash length of 128. 
Both sparse and dense codes are well distributed; i.e. small variance of non zero components per code.

{\bf Evaluation. }
We use several criteria to evaluate the performance of the methods: 
%
%
{\em precision} and {\em recall} (PR) for different Hamming radii, and the {\em F1 score} (their harmonic average);  
{\em mean average precision} at $R$, defined as $mAP = \sum_{n=1}^R P(n) \cdot rel(n)$, where $rel(n)$ is the relevance of the $n$th results (one if relevant and zero otherwise), and $P(n)$ is the precision at $n$ (percentage of relevant results in the first $n$ top-ranked matches); and the 
%
{\em mean precision} (MP), defined as the percentage of correct matches for a fixed number of retrieved elements.   
For the PR curves we use the ranking induced by the Hamming distance between the query and the database samples. 
In case of $r < m$ we considered only the results falling into the Hamming ball of radius $r$. 

%


{\bf CIFAR10} \cite{Krizhevsky09learningmultiple} is a standard set of 60K labeled images belonging to 10 different classes, sampled from the 80M tiny image benchmark \cite{torralba:2008TPAMI}. The images are represented using 384-dimensional GIST descriptors.
Following \cite{liu2012supervised}, we used a training set of 200 images for each class; for testing, we used a disjoint query set of 100 images per class and the remaining 59K images as database.
%

\begin{figure}[t!]
\centering
\includegraphics[width=1\linewidth]{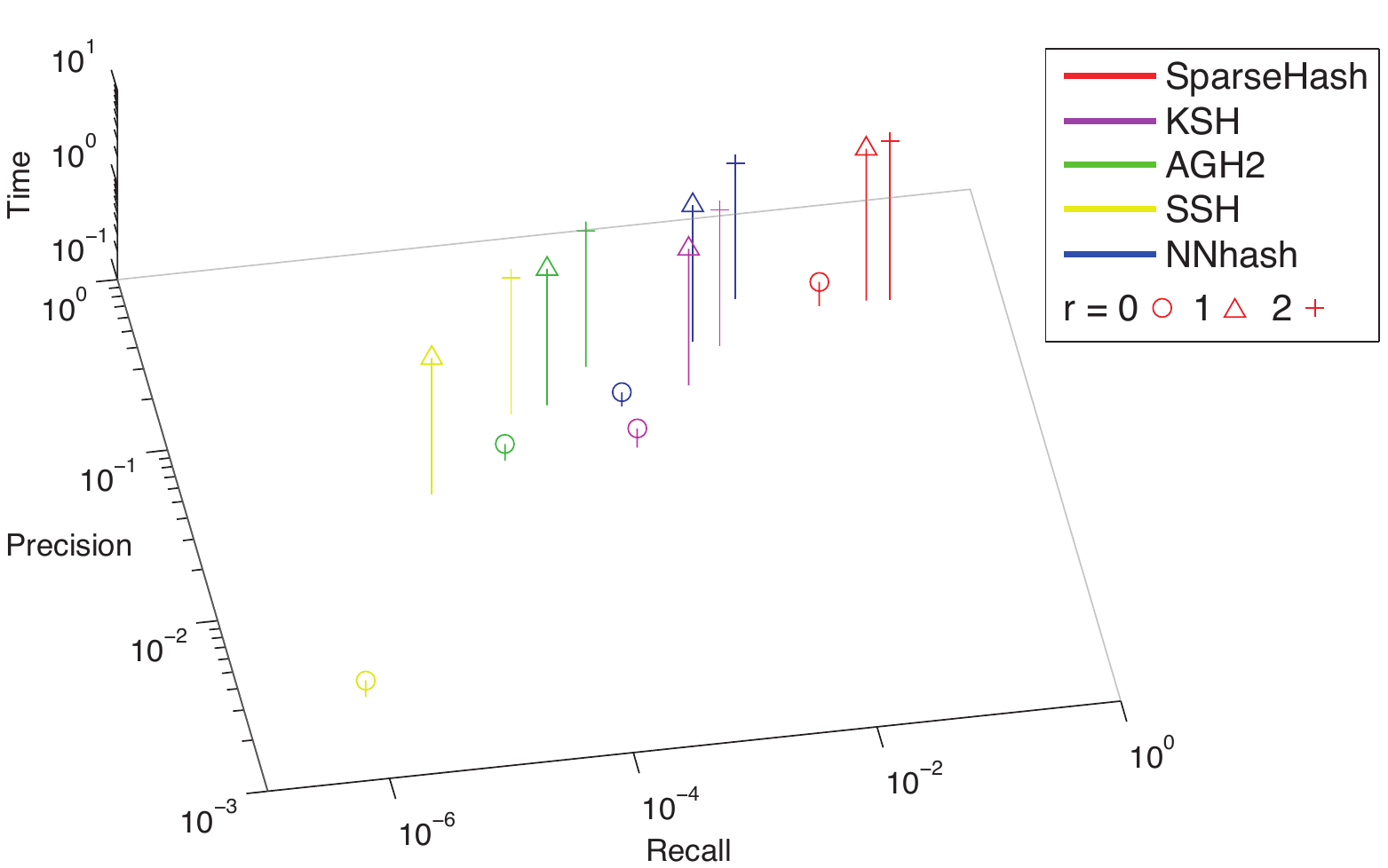}\vspace{-3mm}
\caption{\label{fig:timing}
Tradeoff between recall/precision/query time (in sec) of different hashing methods of length $m=48$ on the CIFAR10 dataset for rejection radii $r=0$ (circle), $1$ (triangle) and $2$ (plus). Retrieval for $r=0,1$ was implemented using LUT; for $r=2$ brute-force search was more efficient. Diffhash produced very low recall and is not shown. 
}\vspace{-4mm}
\end{figure}

\begin{figure}[t!]
\centering
\includegraphics[width=1\linewidth]{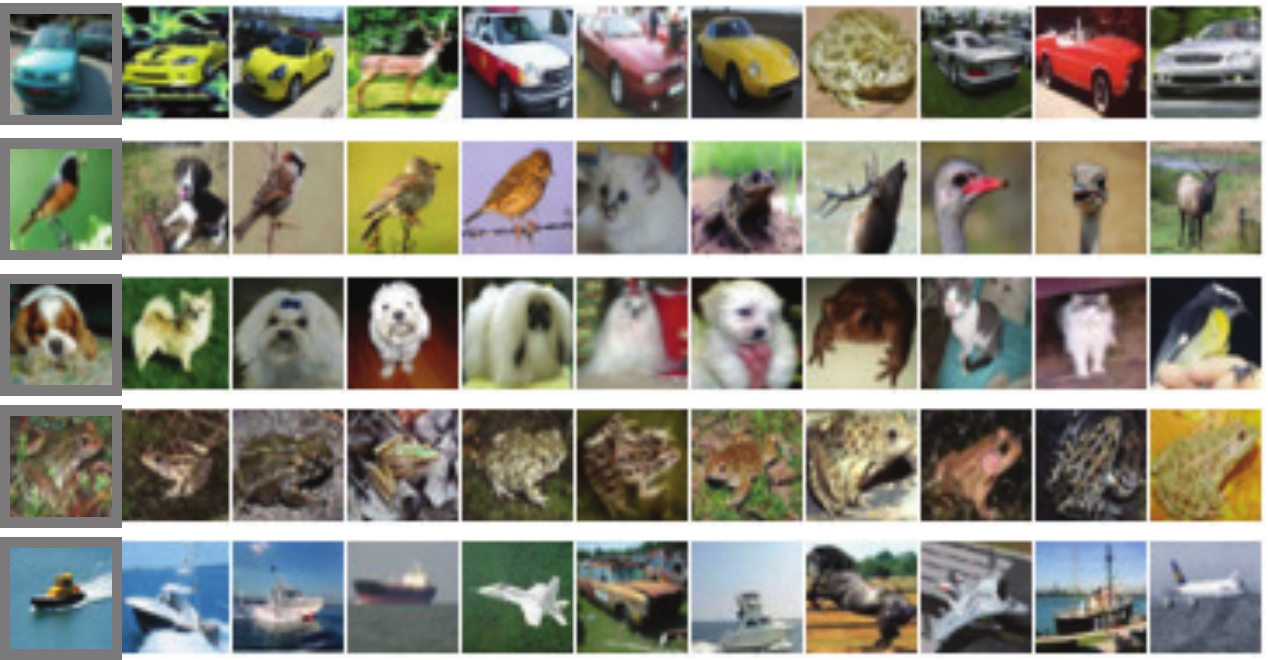}\vspace{-3mm}
\caption{\label{fig:nn-cifar}
Ten nearest neighbors retrieved by SparseHash for five different queries (marked in gray) in the CIFAR10 dataset.}\vspace{-5mm}
\end{figure}

\begin{figure*}[t!]
\centering
\includegraphics[width=1\linewidth]{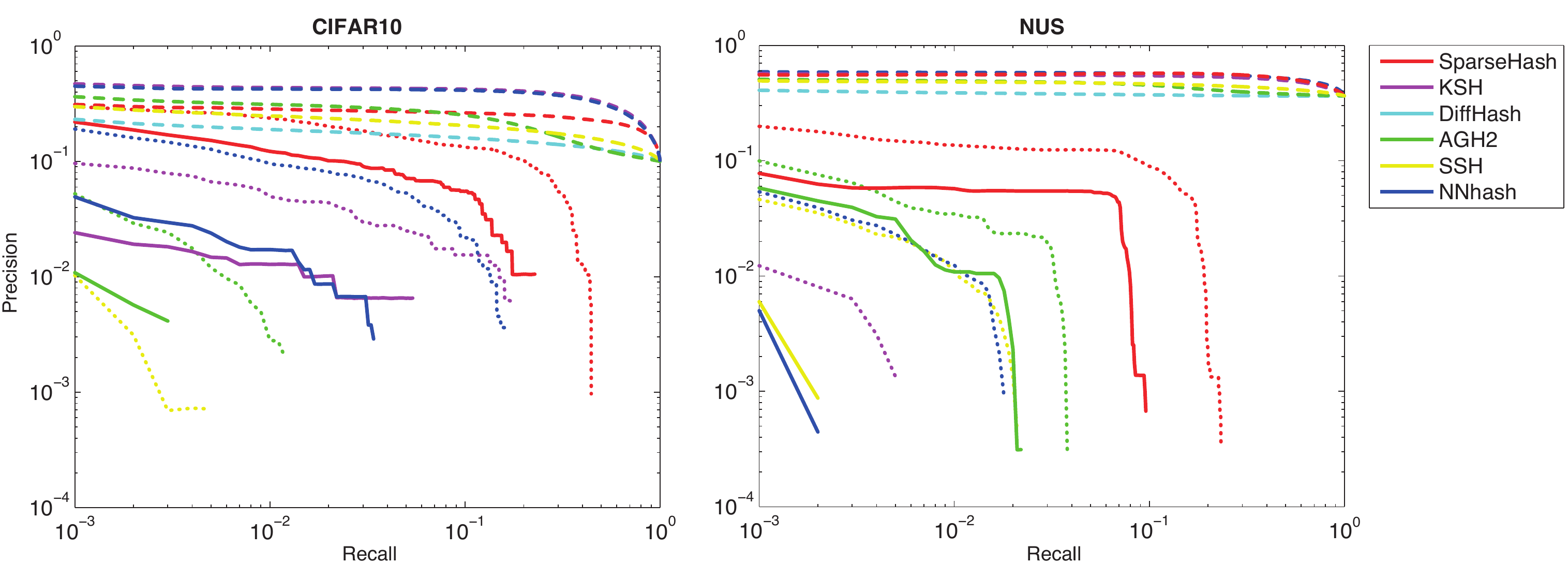}\vspace{-3mm}
\caption{\label{fig:pr-curve}
Precision/recall characteristics of hashing methods using $m=48$ bits for different Hamming radius $r=0$ (solid), $2$ (dotted) and  $m$ (dashed) on CIFAR10 (left) and NUS (right) datasets. 
Some settings result in zero recall and the corresponding curves are not shown. 
While all methods show comparable performance at large $r$, only SparseHash performs well for small values of $r$.
}\vspace{-5mm}
\end{figure*}

\begin{figure*}[t!]
\centering
\includegraphics[width=1\linewidth]{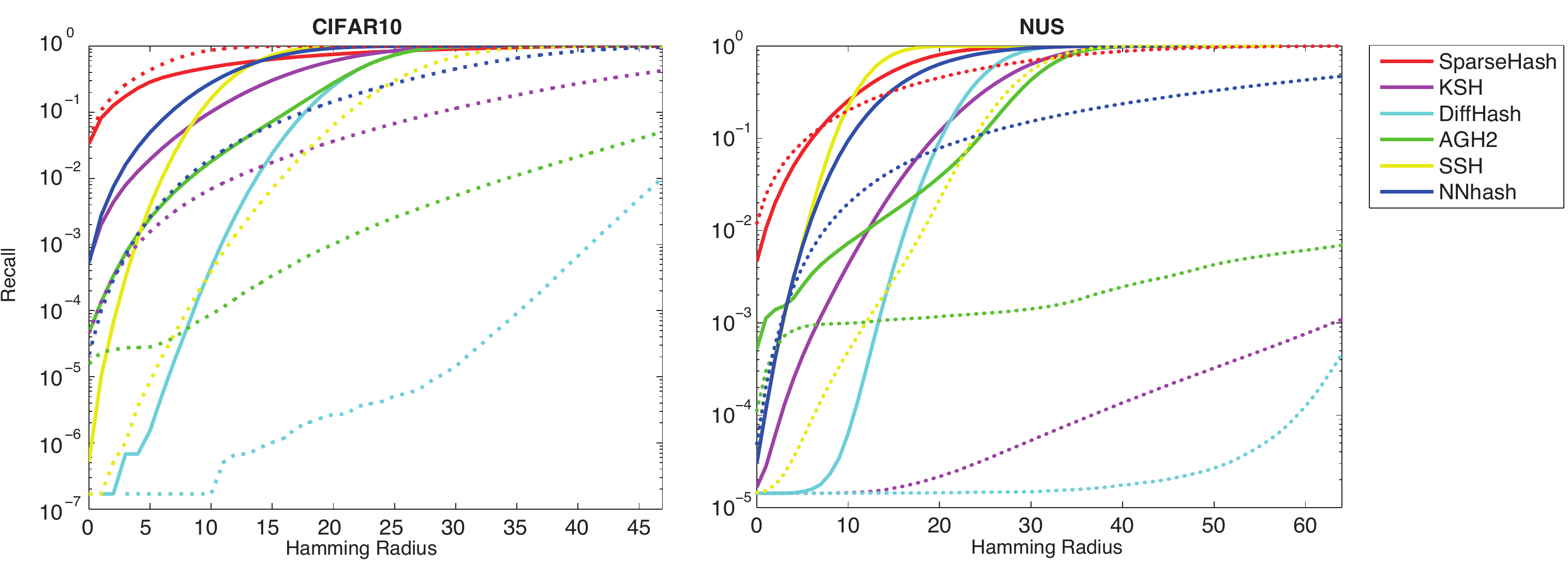}\vspace{-3mm}
\caption{\label{fig:rvsr-curve}
Recall as function of Hamming radius $r$ of hash codes of different length (left: CIFAR10 dataset, solid: $m=48$, dotted: $m=128$; right: NUS dataset, solid: $m=64$, dotted: $m=256$). 
Note the dramatic drop in recall of dense hash methods when increasing code length $m$, while our proposed framework maintains performance.
}\vspace{-3mm}
\end{figure*}

Figure~\ref{fig:nn-cifar} shows examples of nearest neighbors retrieval by SparseHash. Performance of different methods is compared in Table~\ref{tab:cifar10} and Figures~\ref{fig:pr-curve}--\ref{fig:rvsr-curve}.
%
%
%
In Figure~\ref{fig:rvsr-curve} (left), we observe two phenomena: first, the recall of dense hash methods drops significantly with the increase of hash length (as expected from our  analysis in Section~\ref{sec:sparsity}; increasing the hash length is needed for precision performance), while the recall of SparseHash, being dependent on the number of non-zero elements rather than hash length, remains approximately unchanged.  
Second, SparseHash has significantly higher recall at low $r$ compared to other methods.   
This is also evinced in Figure~\ref{fig:timing} where we show the tradeoff between precision, recall and retrieval time for  hashes of length $48$. 
We used an efficient implementation of LUT-based and brute force search and took the fastest among the two; with codes of length $m=48$ on CIFAR10 dataset, for radii $r=0$ and $1$ LUT-based search showed significant speedup, while for $r \ge 2$ brute force search was faster. 
In order to further analyze this behavior, we measured the average number of codes which
are mapped to the same point for each of the methods. Results are reported in Table~\ref{tab:codecollisions}.

\begin{table}[t!]\small
\caption{Total number of unique codes for the entire CIFAR10 dataset and average number of retrieved results
for various Hamming radii search. Hashes of length 48.}
\vspace{-4mm}
\begin{center}
\begin{tabular}{r | c c c c }
& & \multicolumn{3}{c}{\bf Avg. \# of $r$-neighbors } \\
\cline{3-5}
{\bf Method} 	&  {\bf Unique codes} & $r=0$	& $r=1$ & $r=2$ \\
\hline
{\bf KSH	}	& 57368	& 3.95  	& 12.38 	&  27.21 \\
{\bf AGH2 }	& 55863	& 1.42  	& 2.33  	&  4.62 \\
{\bf SSH	}	& 59733 	& 1.01  	& 1.12 	&  1.88 \\
{\bf DH	}	& 59999 	& 1.00  	& 1.00 	&  1.00 \\
{\bf NN	}	& 54259 	& 4.83  	& 20.12 	&  56.70 \\
{\bf Sparse}	& 9828  	& 798.47  & 2034.73	& 3249.86 \\
\end{tabular}
\end{center}
\label{tab:codecollisions}
\vspace{-7mm}
\end{table}

{\bf NUS } \cite{nus-wide-civr09} is a dataset containing 270K annotated images from Flickr. 
Every images is associated with one or more of the different 81 concepts, and is described using a 500-dimensional bag-of-features. 
%
In the training and evaluation, we followed the protocol of \cite{liu2011hashing}: two images were considered as neighbors if they share at least one common concept (only 21 most frequent concepts are considered). 
Testing was done on a query set of 100 images per concept; 
training was performed on 100K pairs of images. 


Performance is shown in Table~\ref{tab:nus_80_256} and Figures~\ref{fig:pr-curve}--\ref{fig:rvsr-curve}; retrieved neighbors are shown in Figure~\ref{fig:nn-nus-5nn}. 
%
We again see behavior consistent with our analysis and SparseHash significant outperforms the other methods. 





\begin{figure*}[th!]
\centering
\includegraphics[width=1\linewidth]{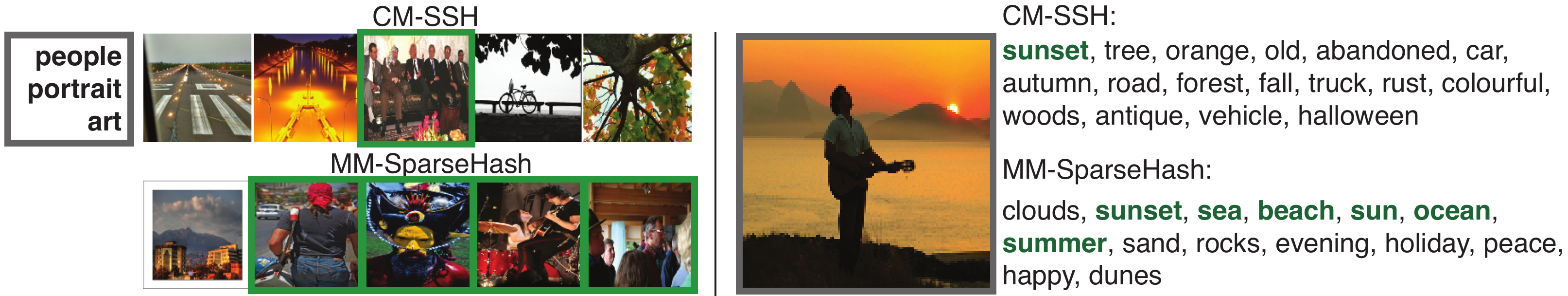}\vspace{-2mm}
\caption{\label{fig:nn-nus}
Cross-modality retrieval on NUS dataset. Left plate: first five Image results to a Tags query (shown in gray) obtained using CM-SSH (top) and MM-SparseHash (bottom). Correct matches are marked in green. Note how many if not most of the results from CM-SSH, considered one of the state-of-the-art techniques in this task, are basically uncorrelated to the query. 
Right plate: union of top five Tags results (right) to an Image query (shown in gray)  obtained using CM-SSH (top) and MM-SparseHash (bottom). Tags matching ground truth tags are shown in green. 
Note how the proposed approach not only detects significantly more matching words, but also the ``non-matching'' ones actually make a lot of sense, indicating that while they were not in the original image tags, underlying connections were learned. Extended version of this figure is shown in supplementary materials. 
}
\vspace{-5mm}
\end{figure*}

{\bf Multi-modal hashing. } 
We repeated the experiment on the NUS dataset with the same indices of positive and negative pairs, adding the \emph{Tags} modality represented as 1K-dimensional bags of words. 
%
The training set contained pairs of similar and dissimilar Images, Tags, and cross-modality Tags-Images pairs. 
%
%
%
We compare our MM-SparseHash to the cross-modal SSH (CM-SSH) method \cite{bronstein2010data}. Results are shown in Table~\ref{tab:mm_nus} and in Figure~\ref{fig:nn-nus}. 
MM-SparseHash significantly outperforms CM-SSH in previously reported state-of-the-art cross-modality (Images-Tags and Tags-Images) retrieval. 
Both methods outperform the $L_2$ baseline for intra-modal (Tags-Tags and Images-Images) retrieval; since the two modalities complement each other, we attribute the improvement to the ability of the model to pick up such correlations. 

%



\input{res_nus.tex}

\input{res_mm.tex}

%% file: res_cifar.tex
\begin{table*}[htdp]
\vspace{-2mm}
\small
\caption{Performance (in \%)  of different hashing methods on the CIFAR10 dataset  with different settings of different length $m$. 
Extended version of this table is shown in supplementary materials. 
\vspace{-2mm}
}
\begin{center}
\begin{tabular}{cc  ccc  | c   c  c c c c c c}
&& &&&		 &   \multicolumn{3}{c}{{\bf Hamming radius} $\leq 2$} & \multicolumn{3}{c}{{\bf Hamming radius} $=0$} \\
\cline{7-12}
 \multicolumn{5}{c|}{{\bf Method}}	& {\bf mAP}  & {\bf Prec.} & {\bf Recall} & {\bf F1} & {\bf Prec.} & {\bf Recall} & {\bf F1} \\
\hline
\multirow{1}{*}{{\bf $L_2$}} 	
&&   && 	
& 17.42 & -- & -- & -- & -- & -- & -- \\
\hline
 & & & & $m$
& & & & & & &  \\
\multirow{2}{*}{{\bf KSH}} 	
&   &&&48			
& 31.10 & 18.22 & 0.86 & 0.44 & 5.39 & 5.6$\times 10^{-2}$ &0.11 \\
&   &&&64			
& 32.49 & 10.86 & 0.13 & 0.26 & 2.49 & 9.6$\times 10^{-3}$ & 1.9$\times 10^{-2}$ \\
%
\hline
\multirow{2}{*}{{\bf AGH1}} 	
&   &&&48		
& 14.55 & 15.95 & 2.8$\times 10^{-2}$ & 1.4$\times 10^{-2}$ & 4.88 & 2.2$\times 10^{-3}$ & 4.4$\times 10^{-3}$ \\
&   &&&64		
& 14.22 & 6.50 & 4.1$\times 10^{-3}$ & 8.1$\times 10^{-3}$ & 3.06 & 1.2$\times 10^{-3}$ & 2.4$\times 10^{-3}$ \\
%
\hline
\multirow{2}{*}{{\bf AGH2}} 	
&   &&&48		
& 15.34 & 17.43 & 7.1$\times 10^{-2}$ & 3.6$\times 10^{-2}$ & 5.44 & 3.5$\times 10^{-3}$ & 6.9$\times 10^{-3}$ \\
&   &&&64
& 14.99 & 7.63 & 7.2$\times 10^{-3}$ & 1.4$\times 10^{-2}$ & 3.61 & 1.4$\times 10^{-3}$ & 2.7$\times 10^{-3}$ \\
%
\hline
\multirow{2}{*}{{\bf SSH}} 	
&   &&&48			
& 15.78 & 9.92 & 6.6$\times 10^{-3}$ & 1.3$\times 10^{-2}$ & 0.30 & 5.1$\times 10^{-5}$ & 1.0$\times 10^{-4}$ \\
&   &&&64
& 17.18 & 1.52 & 3.0$\times 10^{-4}$ & 6.1$\times 10^{-4}$ & 1.0$\times 10^{-3}$ & 1.69$\times 10^{-5}$ & 3.3$\times 10^{-5}$ \\
%
\hline
\multirow{2}{*}{{\bf DH}} 	
&   &&&48	
& 13.13 & 3.0$\times 10^{-3}$ & 1.0$\times 10^{-4}$ & 5.1$\times 10^{-5}$ & 1.0$\times 10^{-3}$ & 1.7$\times 10^{-5}$ & 3.4$\times 10^{-5}$ \\
&   &&&64			
& 13.07 & 1.0$\times 10^{-3}$ & 1.7$\times 10^{-5}$ & 3.3$\times 10^{-5}$ & 0.00 & 0.00 & 0.00 \\
%
\hline
\multirow{2}{*}{{\bf NN}} 	
&&   &&48
& {\bf 30.18} & {\bf 32.69} & 1.45 & 0.74 & 9.47 & 5.2$\times 10^{-2}$ & 0.10 \\
%
%
&   &&&64
& 34.74 & 22.78 & 0.28 & 5.5$\times 10^{-1}$ & 5.70 & 8.8$\times 10^{-3}$ & 1.8$\times 10^{-2}$ \\
%
%
\hline
 &$m$& \hspace{-2mm}$M$\hspace{-2mm} & \hspace{-2mm}$\alpha$ & \hspace{-2mm}$\lambda$\hspace{-2mm} 
& & & & &    \\
\multirow{2}{*}{{\bf Sparse}} &\multirow{2}{*}{48}
& 16 & 0.01 & \hspace{-2mm}0.1 
	& 23.07 & {\bf 32.69} & 1.81 & 0.93 & 16.65 & 5.0$\times 10^{-2}$ & 0.10 \\
%
	%
&& 7 & 0.001 & \hspace{-2mm}0.1 
	& 21.08 & 26.03 & {\bf 17.00} & {\bf 12.56} & 26.65 & {\bf 3.04} & {\bf 5.46}  \\
\cline{2-12}
\multirow{2}{*} &\multirow{2}{*}{64}
& 11 & 0.005 & \hspace{-2mm}0.1 
	& 23.80 & {\bf 31.74} & 6.87 & 11.30 & 31.12 & 0.86 & 1.70 \\
&& 7 & 0.001 & \hspace{-2mm}0.1 
	& 21.29 & 21.41 &{\bf 41.68} & {\bf 28.30} & 25.27 & {\bf 10.17} & {\bf 14.50}  \\

%
%
\hline
\end{tabular}
\end{center}
\label{tab:cifar10}\vspace{-5mm}
\end{table*}%

%% file: res_nus.tex
\begin{table*}[htdp]\small
\vspace{-3mm}
\caption{Performance (in \%) of hashing methods  of different length $m$ on the NUS dataset. 
Extended version of this table is shown in supplementary materials. 
\vspace{-3mm}
}
\begin{center}
\begin{tabular}{c c ccc  | c   c c c c c c c c}
& &&&		& &  & \multicolumn{3}{c}{{\bf Hamming radius} $\leq 2$} & \multicolumn{3}{c}{{\bf Hamming radius} $=0$} \\
\cline{8-13}
 \multicolumn{5}{c|}{{\bf Method}} 	& {\bf mAP@10}  & \hspace{-2mm}{\bf MP@5K}\hspace{-2mm} & {\bf Prec.} & {\bf Recall} & {\bf F1} & {\bf Prec.} & {\bf Recall} & {\bf F1} \\
\hline 
\multirow{1}{*}{} 
{\bf $L_2$}& &  & &  	 & 68.67 & 32.77 & --  & --  & -- & -- & -- & --\\
\hline
 &&  & & $m$
& & & & & & & & \\
\multirow{2}{*}{{\bf KSH}} 
& &   &&64			& 72.85 & 42.74 & 83.80 & 6.1$\times 10^{-3}$ & 1.2$\times 10^{-2}$ & 84.21 & 1.7$\times 10^{-3}$ & 3.3$\times 10^{-3}$\\
& & &&256			& 73.73 & 45.35 & 84.24 & 1.4$\times 10^{-3}$ & 2.9$\times 10^{-3}$ & 84.24 & 1.4$\times 10^{-3}$ & 2.9$\times 10^{-3}$ \\
\hline
\multirow{2}{*}{{\bf AGH1}} 
&&  &&64		& 69.48 & 47.28 & 69.43 & 0.11 & 0.22 & 73.35 & 3.9$\times 10^{-2}$ & 7.9$\times 10^{-2}$ \\
&&  &&256	& 73.86 & 46.68 & 75.90 & 1.5$\times 10^{-2}$ & 2.9$\times 10^{-2}$ & 81.64 & 3.6$\times 10^{-3}$ & 7.1$\times 10^{-3}$ \\
\hline
\multirow{2}{*}{{\bf AGH2}} 
&& &&64		& 68.90 & 47.27 & 68.73 & 0.14 & 0.28 & 72.82 & 5.2$\times 10^{-2}$ & 0.10 \\
&& &&256	& 73.00 & 47.65 & 74.90 & 5.3$\times 10^{-2}$ & 0.11 & 80.45 & 1.1$\times 10^{-2}$ & 2.2$\times 10^{-2}$ \\
\hline
\multirow{2}{*}{{\bf SSH}} 
&& &&64		& 72.17 & 44.79 & 60.06 & 0.12 & 0.24 & 81.73 & 1.1$\times 10^{-2}$ & 2.2$\times 10^{-2}$ \\
&& &&256		& 73.52 & 47.13 & 84.18 & 1.8$\times 10^{-3}$ & 3.5$\times 10^{-3}$ & 84.24 & 1.5$\times 10^{-3}$ & 2.9$\times 10^{-3}$ \\
\hline
\multirow{2}{*}{{\bf DH}} 
&& &&64		& 71.33 & 41.69 & 84.26 & 1.4$\times 10^{-3}$ & 2.9$\times 10^{-3}$ & 84.24 & 1.4$\times 10^{-3}$ & 2.9$\times 10^{-3}$ \\
&& &&256	& 70.73 & 39.02 & 84.24 & 1.4$\times 10^{-3}$ & 2.9$\times 10^{-3}$ & 84.24 & 1.4$\times 10^{-3}$ & 2.9$\times 10^{-3}$ \\
\hline
%
\multirow{2}{*}{{\bf NN}} & &&&\multirow{1}{*}{64}  
	& 76.39 & 59.76 & 75.51 & 1.59 & 3.11 & 81.24 & 0.10 & 0.20 \\
%
%
&&&&\multirow{1}{*}{256}  
	& 78.31 & 61.21 & 83.46 & 5.8$\times 10^{-2}$ & 0.11 & 83.94 & 4.9$\times 10^{-3}$ & 9.8$\times 10^{-3}$ \\
\hline
 &$m$& \hspace{-2mm}$M$\hspace{-2mm} & \hspace{-2mm}$\alpha$ & \hspace{-2mm}$\lambda$\hspace{-2mm} 
& & & & & & &  \\
\multirow{6}{*}{{\bf Sparse}} &\multirow{3}{*}{64}
& 7 & 0.05 & \hspace{-2mm}0.3 
	& 74.17 & 56.08 & 71.67 & {\bf 1.99} & {\bf 3.98} & 81.11 & {\bf 0.46} & {\bf 0.92} \\
 && 7 & 0.05 & \hspace{-2mm}1.0 
	& 74.15 & 51.52 & 69.08 & 0.53 & 1.06 & 81.67 & 0.15 & 0.30\\
&& 16 & \hspace{-2mm}0.005\hspace{-2mm}  & \hspace{-2mm}0.3
& 74.51 & 55.54 & 79.09 & 1.21 & 2.42 & 82.76 & 0.17 & 0.34 \\
\cline{2-13}
&\multirow{3}{*}{256}
& 4 & 0.05 & \hspace{-2mm}1.0 
	& 74.05 & 60.73 & 78.82 & {\bf 3.85} & {\bf 7.34}  & 81.82 & {\bf 1.20} & {\bf 2.37}  \\
 && 4 & 0.05 & \hspace{-2mm}1.0 
	& 74.48 & 59.42 & 81.95 & 1.18 & 2.33  & 83.24 & 0.35 & 0.70 \\
&& 6 & \hspace{-2mm}0.005\hspace{-2mm}  & \hspace{-2mm}0.3 
& 71.73 & 54.76 & 78.34 & 6.10 & 11.30 & 80.85 & 1.02 & 2.01  \\
\hline
\end{tabular} 
\end{center}
\label{tab:nus_80_256}\vspace{-3mm}
\end{table*}%

%% file: res_mm.tex
\begin{table*}[t!]\small
\caption{Performance (in \%) of CM-SSH and MM-SparseHash on the NUS multi-modal dataset with hashes of length 64.}\vspace{-4mm}
\begin{center}
\begin{tabular}{r | ccccc  ccc}
			&  \multicolumn{4}{c}{{\bf mAp10 \%}} &  \multicolumn{4}{c}{{\bf MP@5K \%}} \\
			\cline{2-9}
	{\bf Method} 	& \hspace{-1mm}{\bf Image-Image}\hspace{-1mm} & \hspace{-1mm}{\bf Tag-Tag}\hspace{-1mm} & \hspace{-1mm}{\bf Image-Tag}\hspace{-1mm} & \hspace{-1mm}{\bf Tag-Image}\hspace{+1mm}   & \hspace{+1mm}{\bf  Image-Image}\hspace{-1mm} & \hspace{-1mm}{\bf Tag-Tag}\hspace{-1mm} & \hspace{-1mm}{\bf Image-Tag}\hspace{-1mm} & \hspace{-1mm}{\bf Tag-Image}\hspace{-1mm}  \\
	\hline
$L_2$		& 68.67 & 71.38 & -- & -- & 32.77 & 32.85 & -- & -- \\
\hline
{\bf CM-SSH}		& {\bf 75.19} & 83.05 & 55.55 & 50.43	&	49.69 & 61.60 & 37.05 & 39.13\\
{\bf MM-Sparse}	& 73.79 & {\bf 84.49} & {\bf 61.52} & {\bf 59.52} 	&	{\bf 58.13} & {\bf 66.59} & {\bf 57.35} & {\bf 57.29}\\
\hline
\end{tabular}
\end{center}
\label{tab:mm_nus}\vspace{-5mm}
\end{table*}%

%% file: conclusion.tex
\section{Conclusions}

We presented a new method for learning sparse similarity-preserving hashing functions. 
The hashing is obtained by solving an $\ell_1$ regularized minimization of the aggregate of false positive and false negative rates.
The embedding function is learned by using ISTA-type neural networks \cite{LecunNN}. These networks have a particular architecture very effective for learning discriminative sparse codes. We also show that, once the similarity-preserving hashing problem is stated as training a neural network, it can be straightforwardly extended to the multi-modal setting. While in this work we only used networks with a single layer, more generic embeddings could be learned with this exact framework 
simply by considering multiple layers.

A key contribution of this paper is to show that more accurate nearest neighbor retrieval can be obtained by introducing sparsity into the hashing code. SparseHash can achieve significantly higher recall at the same levels of precision than dense hashing schemes with similar number of degrees of freedom.
At the same time, the sparsity in the hash codes allows retrieving partial collisions at much lower computational complexity than their dense counterparts in a Hamming ball with the same radius. Extensive experimental results backup these claims, showing that the proposed SparseHash framework produces comparable, or superior, results to some of the state-of-the-art methods. 

%% file: supplementary.tex
\begin{figure*}[!h]
\centering
\includegraphics[width=1\linewidth]{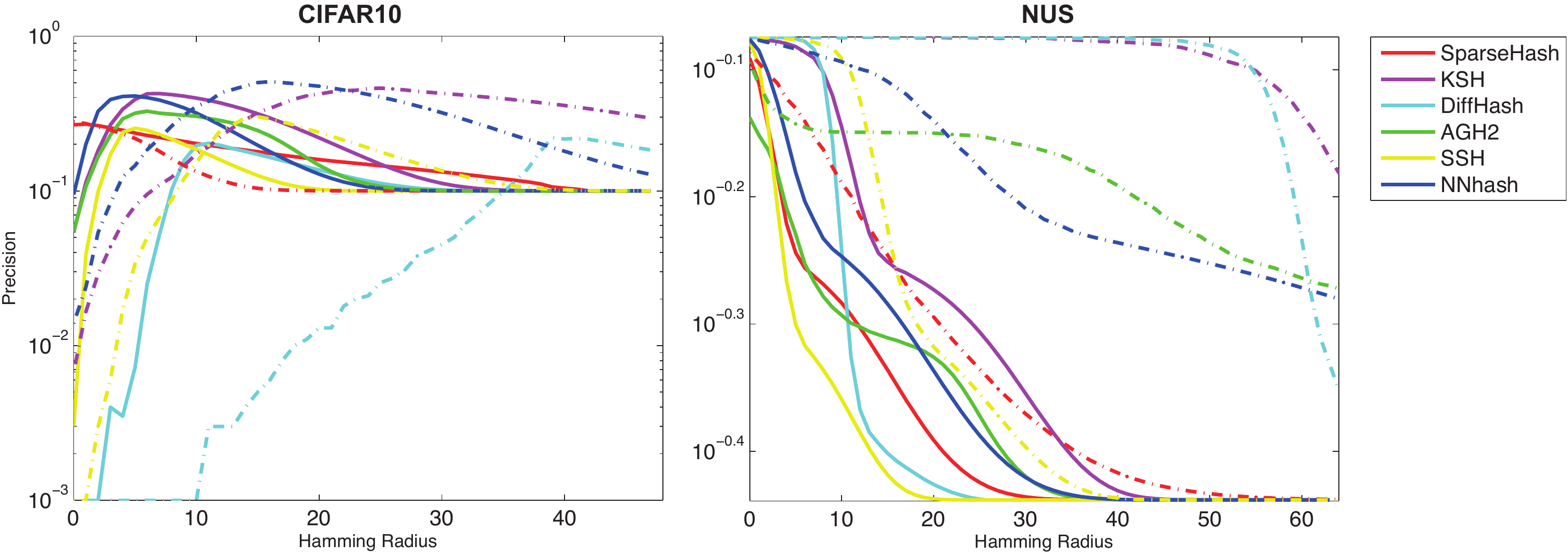}
\caption{\label{fig:pvsr-curve}
Precision as function of Hamming radius $r$ of hash codes of different length (left: CIFAR10 dataset, solid: $m=48$, dotted: $m=128$; right: NUS dataset, solid: $m=64$, dotted: $m=256$). 
}
\end{figure*}

\begin{figure}[!h]
\centering
\includegraphics[width=1\linewidth]{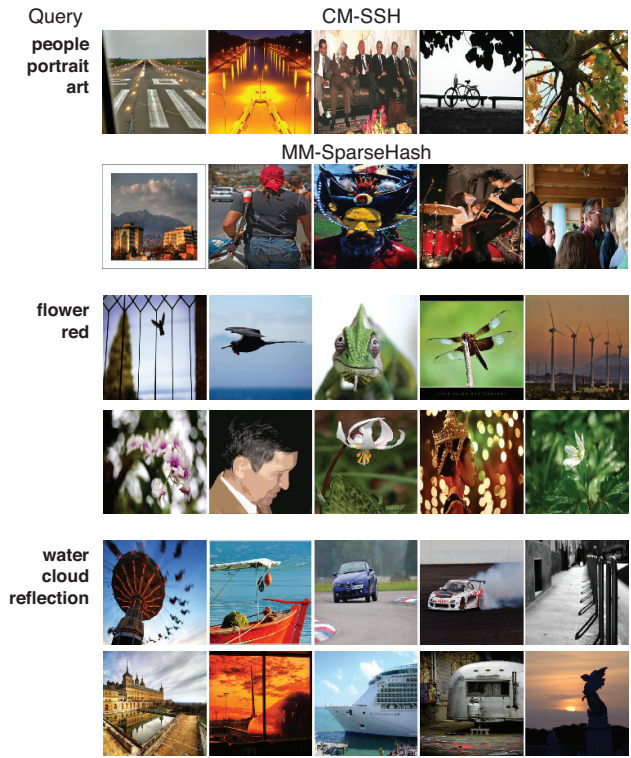}\\
\caption{\label{fig:nn-nus}
Cross-modality retrieval: first five Image results to three Tags queries (left)  obtained using CM-SSH (odd rows) and MM-SparseHash (even rows) in the multimodal NUS dataset. 
Note how many if not most of the results from CM-SSH, considered one of the state-of-the-art techniques in this task, are basically uncorrelated to the query.} 
\end{figure}


\begin{table*}[!h]
\small
\caption{Performance (in \%)  of different hashing methods on the CIFAR10 dataset  with different settings of different length $m$. 
}
\begin{center}
\begin{tabular}{cc  ccc  | c   c  c c c c c c}
&& &&&		 &   \multicolumn{3}{c}{{\bf Hamming radius} $\leq 2$} & \multicolumn{3}{c}{{\bf Hamming radius} $=0$} \\
\cline{7-12}
 \multicolumn{5}{c|}{{\bf Method}}	& {\bf mAP}  & {\bf Prec.} & {\bf Recall} & {\bf F1} & {\bf Prec.} & {\bf Recall} & {\bf F1} \\
\hline
\multirow{1}{*}{{\bf $L_2$}} 	
&&   && 	
& 17.42 & -- & -- & -- & -- & -- & -- \\
\hline
 & & & & $m$
& & & & & & &  \\
\multirow{3}{*}{{\bf KSH}} 	
&   &&&48			
& 31.10 & 18.22 & 0.44 & 0.86 & 5.39 & 5.6$\times 10^{-2}$ &0.11 \\
&   &&&64			
& 32.49 & 10.86 & 0.13 & 0.26 & 2.49 & 9.6$\times 10^{-3}$ & 1.9$\times 10^{-2}$ \\
&   &&&128			
& 33.50 & 2.91 & 3.3$\times 10^{-3}$ & 6.5$\times 10^{-2}$ & 0.67 & 4.5$\times 10^{-3}$ & 8.9$\times 10^{-3}$ \\
\hline
\multirow{3}{*}{{\bf AGH1}} 	
&   &&&48		
& 14.55 & 15.95 & 1.4$\times 10^{-2}$ & 2.8$\times 10^{-2}$ & 4.88 & 2.2$\times 10^{-3}$ & 4.4$\times 10^{-3}$ \\
&   &&&64		
& 14.22 & 6.50 & 4.1$\times 10^{-3}$ & 8.1$\times 10^{-3}$ & 3.06 & 1.2$\times 10^{-3}$ & 2.4$\times 10^{-3}$ \\
&   &&&128			
& 13.53 & 2.89 & 1.1$\times 10^{-3}$ & 2.2$\times 10^{-3}$ & 1.58 & 3.4$\times 10^{-4}$ & 6.8$\times 10^{-4}$ \\
\hline
\multirow{3}{*}{{\bf AGH2}} 	
&   &&&48		
& 15.34 & 17.43 & 3.6$\times 10^{-2}$ & 7.1$\times 10^{-2}$ & 5.44 & 3.5$\times 10^{-3}$ & 6.9$\times 10^{-3}$ \\
&   &&&64
& 14.99 & 7.63 & 7.2$\times 10^{-3}$ & 1.4$\times 10^{-2}$ & 3.61 & 1.4$\times 10^{-3}$ & 2.7$\times 10^{-3}$ \\
&   &&&128			
& 14.38 & 3.78 & 1.6$\times 10^{-3}$ & 3.2$\times 10^{-3}$ & 1.43 & 3.9$\times 10^{-4}$ & 7.8$\times 10^{-4}$ \\
\hline
\multirow{3}{*}{{\bf SSH}} 	
&   &&&48			
& 15.78 & 9.92 & 6.6$\times 10^{-3}$ & 1.3$\times 10^{-2}$ & 0.30 & 5.1$\times 10^{-5}$ & 1.0$\times 10^{-4}$ \\
&   &&&64
& 17.18 & 1.52 & 3.1$\times 10^{-4}$ & 6.1$\times 10^{-4}$ & 1.0$\times 10^{-3}$ & 1.7$\times 10^{-5}$ & 3.3$\times 10^{-5}$ \\
&   &&&128			
& 17.20 & 0.30 & 5.1$\times 10^{-5}$ & 1.0$\times 10^{-4}$ & 0.10 & 1.7$\times 10^{-5}$ & 3.4$\times 10^{-5}$ \\
\hline
\multirow{3}{*}{{\bf DH}} 	
&   &&&48	
& 13.13 & 3.0$\times 10^{-3}$ & 5.1$\times 10^{-5}$ & 1.0$\times 10^{-4}$ & 1.0$\times 10^{-3}$ & 1.7$\times 10^{-5}$ & 3.4$\times 10^{-5}$ \\
&   &&&64			
& 13.07 & 1.0$\times 10^{-3}$ & 1.7$\times 10^{-5}$ & 3.3$\times 10^{-5}$ & 0.00 & 0.00 & 0.00 \\
&   &&&128			
& 13.12 & 0.00 & 0.00 & 0.00 & 0.00 & 0.00 & 0.00 \\
\hline
\multirow{3}{*}{{\bf NN}} 	
&&   &&48
& {\bf 30.18} & {\bf 32.69} & 0.74 & 1.45 & 9.47 & 5.2$\times 10^{-2}$ & 0.10 \\
&   &&&64
& 34.74 & 22.78 & 0.28 & 5.5$\times 10^{-1}$ & 5.70 & 8.8$\times 10^{-3}$ & 1.8$\times 10^{-2}$ \\
&   &&&128			
& 37.89 & 5.38 & 2.9$\times 10^{-2}$ & 5.7$\times 10^{-2}$ & 1.39 & 2.2$\times 10^{-3}$ & 4.4$\times 10^{-3}$ \\
\hline
 &$m$& \hspace{-2mm}$M$\hspace{-2mm} & \hspace{-2mm}$\alpha$ & \hspace{-2mm}$\lambda$\hspace{-2mm} 
& & & & &    \\
\multirow{5}{*}{{\bf Sparse}} &\multirow{2}{*}{48}
& 16 & 0.01 & \hspace{-2mm}0.1 
	& 23.07 & {\bf 32.69} & 0.93 & 1.81 & 16.65 & 5.0$\times 10^{-2}$ & 0.10 \\
%
	%
&& 7 & 0.001 & \hspace{-2mm}0.1 
	& 21.08 & 26.03 &{\bf 12.56} & {\bf 17.00} & 26.65 & {\bf 3.04} & {\bf 5.46}  \\
\cline{2-12}
\multirow{2}{*} &\multirow{2}{*}{64}
& 11 & 0.005 & \hspace{-2mm}0.1 
	& 23.80 & {\bf 31.74} & 6.87 & 11.30 & 31.12 & 0.86 & 1.70 \\
&& 7 & 0.001 & \hspace{-2mm}0.1 
	& 21.29 & 21.41 &{\bf 41.68} & {\bf 28.30} & 25.27 & {\bf 10.17} & {\bf 14.50}  \\
\cline{2-12}
 &\multirow{1}{*}{128}
& 16 & 0 & \hspace{-2mm}0.1 
	& 21.97 & 25.94 & {\bf 18.11} & {\bf 21.30} & 27.99 & {\bf 3.81} & {\bf 6.71}  \\
\hline
\end{tabular}
\end{center}
\label{tab:cifar10} 
\end{table*}%


\begin{table*}[!h]
\small
\caption{Performance (in \%) of hashing methods  of different length $m$ on the NUS dataset. 
Comparable degrees of freedom are 256-bit SparseHash and 80-bit dense hashes. 
}
\begin{center}
\begin{tabular}{c c ccc  | c   c c c c c c c c}
& &&&		& &  & \multicolumn{3}{c}{{\bf Hamming radius} $\leq 2$} & \multicolumn{3}{c}{{\bf Hamming radius} $=0$} \\
\cline{8-13}
 \multicolumn{5}{c|}{{\bf Method}} 	& {\bf mAP@10}  & \hspace{-2mm}{\bf MP@5K}\hspace{-2mm} & {\bf Prec.} & {\bf Recall} & {\bf F1} & {\bf Prec.} & {\bf Recall} & {\bf F1} \\

\hline 
\multirow{1}{*}{} 
{\bf $L_2$}& &  & &  	 & 68.67 & 32.77 &  &  &  &  &  & \\
\hline
 &&  & & $m$
& & & & & & & & \\
\multirow{2}{*}{{\bf KSH}} 
& &   &&64			& 72.85 & 42.74 & 83.80 & 6.1$\times 10^{-3}$ & 1.2$\times 10^{-2}$ & 84.21 & 1.7$\times 10^{-3}$ & 3.3$\times 10^{-3}$\\
& &  &&80		& 72.76 & 43.32 & 84.21 & 1.8$\times 10^{-3}$ & 3.6$\times 10^{-3}$ & 84.23 & 1.4$\times 10^{-3}$ & 2.9$\times 10^{-3}$ \\
& & &&256			& 73.73 & 45.35 & 84.24 & 1.4$\times 10^{-3}$ & 2.9$\times 10^{-3}$ & 84.24 & 1.4$\times 10^{-3}$ & 2.9$\times 10^{-3}$ \\
\hline
\multirow{2}{*}{{\bf AGH1}} 
&&  &&64		& 69.48 & 47.28 & 69.43 & 0.11 & 0.22 & 73.35 & 3.9$\times 10^{-2}$ & 7.9$\times 10^{-2}$ \\
&&  &&80		& 69.62 & 47.23 & 71.15 & 7.5$\times 10^{-2}$ & 0.15 & 74.14 & 2.5$\times 10^{-3}$ & 5.1$\times 10^{-2}$ \\
&&  &&256	& 73.86 & 46.68 & 75.90 & 1.5$\times 10^{-2}$ & 2.9$\times 10^{-2}$ & 81.64 & 3.6$\times 10^{-3}$ & 7.1$\times 10^{-3}$ \\
\hline
\multirow{2}{*}{{\bf AGH2}} 
&& &&64		& 68.90 & 47.27 & 68.73 & 0.14 & 0.28 & 72.82 & 5.2$\times 10^{-2}$ & 0.10 \\
&& &&80	& 69.73 & 47.32 & 70.57 & 0.12 & 0.24 & 73.85 & 4.2$\times 10^{-2}$ & 8.3$\times 10^{-2}$ \\
&& &&256	& 73.00 & 47.65 & 74.90 & 5.3$\times 10^{-2}$ & 0.11 & 80.45 & 1.1$\times 10^{-2}$ & 2.2$\times 10^{-2}$ \\
\hline
\multirow{2}{*}{{\bf SSH}} 
&& &&64		& 72.17 & 44.79 & 60.06 & 0.12 & 0.24 & 81.73 & 1.1$\times 10^{-2}$ & 2.2$\times 10^{-2}$ \\
&& &&80			& 72.58 & 46.96 & 83.96 & 1.9$\times 10^{-3}$ & 3.9$\times 10^{-3}$ & 80.91 & 1.3$\times 10^{-2}$ & 2.6$\times 10^{-2}$ \\
&& &&256		& 73.52 & 47.13 & 84.18 & 1.8$\times 10^{-3}$ & 3.5$\times 10^{-3}$ & 84.24 & 1.5$\times 10^{-3}$ & 2.9$\times 10^{-3}$ \\
\hline
\multirow{2}{*}{{\bf DH}} 
&& &&64		& 71.33 & 41.69 & 84.26 & 1.4$\times 10^{-3}$ & 2.9$\times 10^{-3}$ & 84.24 & 1.4$\times 10^{-3}$ & 2.9$\times 10^{-3}$ \\
&& &&80	& 70.34 & 37.75 & 84.24 & 4.9$\times 10^{-3}$ & 9.8$\times 10^{-3}$ & 84.24 & 4.9$\times 10^{-3}$ & 9.8$\times 10^{-3}$ \\
&& &&256	& 70.73 & 39.02 & 84.24 & 1.4$\times 10^{-3}$ & 2.9$\times 10^{-3}$ & 84.24 & 1.4$\times 10^{-3}$ & 2.9$\times 10^{-3}$ \\
\hline
%
\multirow{2}{*}{{\bf NN}} & &&&\multirow{1}{*}{64}  
	& 76.39 & 59.76 & 75.51 & 1.59 & 3.11 & 81.24 & 0.10 & 0.20 \\
%
&&&&\multirow{1}{*}{80}  
	& 75.51 & 59.59 & 77.17 & 2.02 & 3.94  & 81.89 & 0.24 & 0.48 \\
%
&&&&\multirow{1}{*}{256}  
	& 78.31 & 61.21 & 83.46 & 5.8$\times 10^{-2}$ & 0.11 & 83.94 & 4.9$\times 10^{-3}$ & 9.8$\times 10^{-3}$ \\
\hline
 &$m$& \hspace{-2mm}$M$\hspace{-2mm} & \hspace{-2mm}$\alpha$ & \hspace{-2mm}$\lambda$\hspace{-2mm} 
& & & & & & &  \\
\multirow{6}{*}{{\bf Sparse}} &\multirow{3}{*}{64}
& 7 & 0.05 & \hspace{-2mm}0.3 
	& 74.17 & 56.08 & 71.67 & {\bf 1.99} & {\bf 3.98} & 81.11 & {\bf 0.46} & {\bf 0.92} \\
 && 7 & 0.05 & \hspace{-2mm}1.0 
	& 74.15 & 51.52 & 69.08 & 0.53 & 1.06 & 81.67 & 0.15 & 0.30\\
&& 16 & \hspace{-2mm}0.005\hspace{-2mm}  & \hspace{-2mm}0.3
& 74.51 & 55.54 & 79.09 & 1.21 & 2.42 & 82.76 & 0.17 & 0.34 \\
\cline{2-13}
&\multirow{3}{*}{256}
& 4 & 0.05 & \hspace{-2mm}1.0 
	& 74.05 & 60.73 & 78.82 & {\bf 3.85} & {\bf 7.34}  & 81.82 & {\bf 1.20} & {\bf 2.37}  \\
 && 4 & 0.05 & \hspace{-2mm}1.0 
	& 74.48 & 59.42 & 81.95 & 1.18 & 2.33  & 83.24 & 0.35 & 0.70 \\
&& 6 & \hspace{-2mm}0.005\hspace{-2mm}  & \hspace{-2mm}0.3 
& 71.73 & 54.76 & 78.34 & 6.10 & 11.30 & 80.85 & 1.02 & 2.01  \\
\hline
\end{tabular} 
\end{center}
\label{tab:nus_80_256} 
\end{table*}%

\begin{figure}[!htbp]
\centering
\includegraphics[width=1.\linewidth]{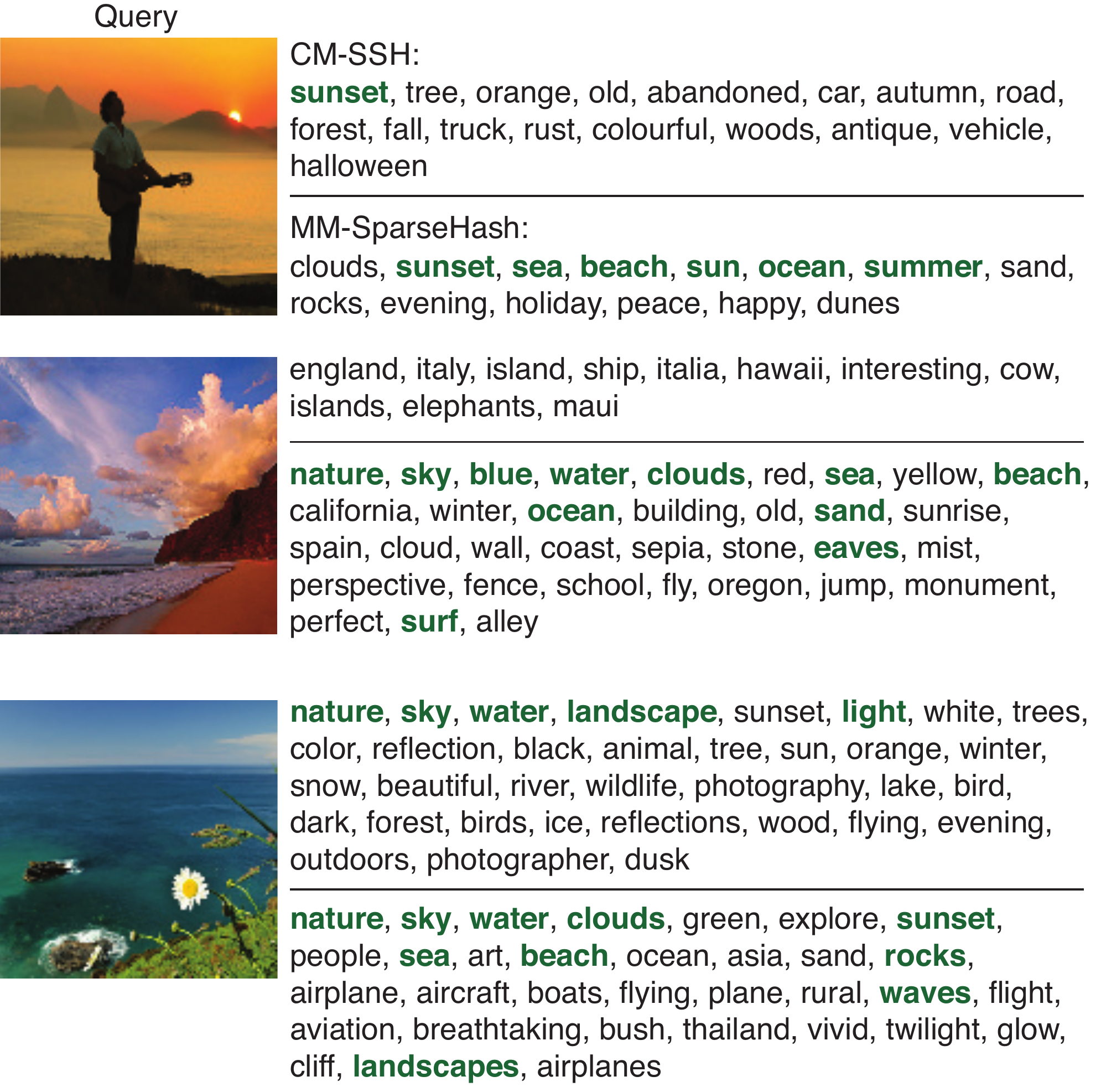}\\
\caption{\label{fig:nn-nus2}
Cross-modality retrieval: union of top five Tags results (right) to three Image queries (left)  obtained using CM-SSH (odd rows) and MM-SparseHash (even rows) in the multimodal NUS dataset. Tags matching ground truth tags showed in green. 
Note how the proposed approach not only detects significantly more matching words, but also the ``non-matching'' ones actually make a lot of sense, indicating that while they were not in the original image tags, underlying connections were learned.} 
\end{figure}


%% file: SparseHash.bbl
\begin{thebibliography}{39}
\providecommand{\natexlab}[1]{#1}
\providecommand{\url}[1]{\texttt{#1}}
\expandafter\ifx\csname urlstyle\endcsname\relax
  \providecommand{\doi}[1]{doi: #1}\else
  \providecommand{\doi}{doi: \begingroup \urlstyle{rm}\Url}\fi

\bibitem[Belkin \& Niyogi(2003)Belkin and Niyogi]{belkin2003laplacian}
Belkin, M. and Niyogi, P.
\newblock Laplacian eigenmaps for dimensionality reduction and data
  representation.
\newblock \emph{Neural computation}, 15\penalty0 (6):\penalty0 1373--1396,
  2003.

\bibitem[Bronstein et~al.(2010)]{bronstein2010data}
Bronstein, M.~M. et~al.
\newblock Data fusion through cross-modality metric learning using
  similarity-sensitive hashing.
\newblock In \emph{Proc. CVPR}, 2010.

\bibitem[Chua et~al.(2009)]{nus-wide-civr09}
Chua, T.-S. et~al.
\newblock {NUS-WIDE}: A real-world web image database from national university
  of {S}ingapore.
\newblock In \emph{Proc. CIVR}, 2009.

\bibitem[Coifman \& Lafon(2006)Coifman and Lafon]{coifman2006diffusion}
Coifman, R.~R. and Lafon, S.
\newblock Diffusion maps.
\newblock \emph{App. Comp. Harmonic Analysis}, 21\penalty0 (1):\penalty0 5--30,
  2006.

\bibitem[Daubechies et~al.(2004)Daubechies, Defrise, and
  De~Mol]{daubechies2004iterative}
Daubechies, I., Defrise, M., and De~Mol, C.
\newblock An iterative thresholding algorithm for linear inverse problems with
  a sparsity constraint.
\newblock \emph{Comm. Pure and App. Math.}, 57\penalty0 (11):\penalty0
  1413--1457, 2004.

\bibitem[Davis et~al.(2007)]{davis2007information}
Davis et~al.
\newblock Information-theoretic metric learning.
\newblock In \emph{Proc. ICML}, 2007.

\bibitem[Gionis et~al.(1999)Gionis, Indyk, and Motwani]{gionis1999similarity}
Gionis, A., Indyk, P., and Motwani, R.
\newblock Similarity search in high dimensions via hashing.
\newblock In \emph{Proc. VLDB}, 1999.

\bibitem[Goemans \& Williamson(1995)Goemans and
  Williamson]{goemans1995improved}
Goemans, M. and Williamson, D.
\newblock Improved approximation algorithms for maximum cut and satisfiability
  problems using semidefinite programming.
\newblock \emph{J. ACM}, 42\penalty0 (6):\penalty0 1115--1145, 1995.

\bibitem[Gong \& Lazebnik(2011)Gong and Lazebnik]{gong2011iterative}
Gong, Y. and Lazebnik, S.
\newblock Iterative quantization: A procrustean approach to learning binary
  codes.
\newblock In \emph{Proc. CVPR}, 2011.

\bibitem[Gong et~al.(2012)]{gong2012angular}
Gong, Y. et~al.
\newblock Angular quantization-based binary codes for fast similarity search.
\newblock In \emph{Proc. NIPS}, 2012.

\bibitem[Grauman \& Fergus(2013)Grauman and Fergus]{grauman2013learning}
Grauman, K. and Fergus, R.
\newblock Learning binary hash codes for large-scale image search.
\newblock In \emph{Machine Learning for Computer Vision}, pp.\  49--87.
  Springer, 2013.

\bibitem[Gregor \& LeCun(2010)Gregor and LeCun]{LecunNN}
Gregor, K. and LeCun, Y.
\newblock Learning fast approximations of sparse coding.
\newblock In \emph{ICML}, 2010.

\bibitem[Hadsell et~al.(2006)Hadsell, Chopra, and
  LeCun]{hadsell-chopra-lecun-06}
Hadsell, R., Chopra, S., and LeCun, Y.
\newblock Dimensionality reduction by learning an invariant mapping.
\newblock In \emph{Proc. CVPR}, 2006.

\bibitem[Johnson \& Wichern(2002)Johnson and Wichern]{johnson2002applied}
Johnson, R.~A. and Wichern, D.~W.
\newblock \emph{Applied multivariate statistical analysis}, volume~4.
\newblock Prentice Hall, 2002.

\bibitem[Korman \& Avidan(2011)Korman and Avidan]{korman2011coherency}
Korman, S. and Avidan, S.
\newblock Coherency sensitive hashing.
\newblock In \emph{Proc. ICCV}, 2011.

\bibitem[Krizhevsky(2009)]{Krizhevsky09learningmultiple}
Krizhevsky, Alex.
\newblock Learning multiple layers of features from tiny images.
\newblock Technical report, 2009.

\bibitem[Kulis \& Darrell(2009)Kulis and Darrell]{kulis2009learning}
Kulis, B. and Darrell, T.
\newblock Learning to hash with binary reconstructive embeddings.
\newblock In \emph{Proc. NIPS}, 2009.

\bibitem[LeCun(1985)]{LeCun:85}
LeCun, Y.
\newblock Une proc\'{e}dure d'apprentissage pour r\'{e}seau \`{a} seuil
  asym\'{e}trique.
\newblock \emph{Proceedings of Cognitiva 85, Paris}, pp.\  599--604, 1985.

\bibitem[Liu et~al.(2011)]{liu2011hashing}
Liu, W. et~al.
\newblock Hashing with graphs.
\newblock In \emph{Proc. ICML}, 2011.

\bibitem[Liu et~al.(2012)]{liu2012supervised}
Liu, Wei et~al.
\newblock Supervised hashing with kernels.
\newblock In \emph{Proc. CVPR}, 2012.

\bibitem[Masci et~al.(2011)]{masci2011descriptor}
Masci, J. et~al.
\newblock Descriptor learning for omnidirectional image matching.
\newblock Technical Report arXiv:1112.6291, 2011.

\bibitem[McFee \& Lanckriet(2009)McFee and Lanckriet]{mcfee2009partial}
McFee, B. and Lanckriet, G. R.~G.
\newblock Partial order embedding with multiple kernels.
\newblock In \emph{Proc. ICML}, 2009.

\bibitem[Mika et~al.(1999)]{mika1999fisher}
Mika, S. et~al.
\newblock Fisher discriminant analysis with kernels.
\newblock In \emph{Proc. Neural Networks for Signal Processing}, 1999.

\bibitem[Norouzi \& Fleet(2011)Norouzi and Fleet]{norouzi2011minimal}
Norouzi, M. and Fleet, D.
\newblock Minimal loss hashing for compact binary codes.
\newblock In \emph{Proc. ICML}, 2011.

\bibitem[Norouzi et~al.(2012)Norouzi, Fleet, and
  Salakhutdinov]{norouzi2012hamming}
Norouzi, M., Fleet, D., and Salakhutdinov, R.
\newblock Hamming distance metric learning.
\newblock In \emph{Proc. NIPS}, 2012.

\bibitem[Roweis \& Saul(2000)Roweis and Saul]{roweis2000nonlinear}
Roweis, S.~T. and Saul, L.~K.
\newblock Nonlinear dimensionality reduction by locally linear embedding.
\newblock \emph{Science}, 290\penalty0 (5500):\penalty0 2323, 2000.

\bibitem[Schoelkopf et~al.(1997)Schoelkopf, Smola, and
  Mueller]{scholkopf1997kernel}
Schoelkopf, B., Smola, A., and Mueller, K.~R.
\newblock Kernel principal component analysis.
\newblock \emph{Artificial Neural Networks}, pp.\  583--588, 1997.

\bibitem[Shakhnarovich et~al.(2003)Shakhnarovich, Viola, and
  Darrell]{shakhnarovich2003fast}
Shakhnarovich, G., Viola, P., and Darrell, T.
\newblock Fast pose estimation with parameter-sensitive hashing.
\newblock In \emph{Proc. CVPR}, 2003.

\bibitem[Shen et~al.(2009)]{shen2009positive}
Shen, C. et~al.
\newblock Positive semidefinite metric learning with boosting.
\newblock In \emph{Proc. NIPS}, 2009.

\bibitem[Sprechmann et~al.(2012)Sprechmann, Bronstein, and Sapiro]{pami2012}
Sprechmann, P., Bronstein, A.~M., and Sapiro, G.
\newblock Learning efficient sparse and low rank models.
\newblock Technical Report arXiv:1010.3467, 2012.

\bibitem[Strecha et~al.(2012)]{strecha2010ldahash}
Strecha, C. et~al.
\newblock {LDAHash}: Improved matching with smaller descriptors.
\newblock \emph{PAMI}, 34\penalty0 (1):\penalty0 66--78, 2012.

\bibitem[Taylor et~al.(2011)]{Taylor11}
Taylor, G.~W. et~al.
\newblock Learning invariance through imitation.
\newblock In \emph{Proc. CVPR}, 2011.

\bibitem[Torralba et~al.(2008{\natexlab{a}})Torralba, Fergus, and
  Freeman]{torralba:2008TPAMI}
Torralba, A., Fergus, R., and Freeman, W.~T.
\newblock 80 million tiny images: A large data set for nonparametric object and
  scene recognition.
\newblock \emph{PAMI}, 30\penalty0 (11):\penalty0 1958--1970,
  2008{\natexlab{a}}.

\bibitem[Torralba et~al.(2008{\natexlab{b}})Torralba, Fergus, and
  Weiss]{torralba2008small}
Torralba, A., Fergus, R., and Weiss, Y.
\newblock Small codes and large image databases for recognition.
\newblock In \emph{Proc. CVPR}, 2008{\natexlab{b}}.

\bibitem[Wang et~al.(2010)Wang, Kumar, and Chang]{wang2010sequential}
Wang, J., Kumar, S., and Chang, S.-F.
\newblock Sequential projection learning for hashing with compact codes.
\newblock In \emph{Proc. ICML}, 2010.

\bibitem[Weinberger \& Saul(2009)Weinberger and Saul]{weinberger2009distance}
Weinberger, K.~Q. and Saul, L.~K.
\newblock Distance metric learning for large margin nearest neighbor
  classification.
\newblock \emph{JMLR}, 10:\penalty0 207--244, 2009.

\bibitem[Weiss et~al.(2008)Weiss, Torralba, and Fergus]{weiss2008spectral}
Weiss, Y., Torralba, A., and Fergus, R.
\newblock Spectral hashing.
\newblock In \emph{Proc. NIPS}, 2008.

\bibitem[Xing et~al.(2002)]{xing2002distance}
Xing, E.~P. et~al.
\newblock Distance metric learning with application to clustering with
  side-information.
\newblock In \emph{Proc. NIPS}, 2002.

\bibitem[Yagnik et~al.(2011)]{yagnik2011power}
Yagnik, J. et~al.
\newblock The power of comparative reasoning.
\newblock In \emph{Proc. CVPR}, 2011.

\end{thebibliography}
